# Monotonicity in Bayesian Networks

**Linda C. van der Gaag**, **Hans L. Bodlaender**, and **Ad Feelders**
Institute of Information and Computing Sciences, Utrecht University,
P.O. Box 80.089, 3508 TB Utrecht, the Netherlands.
e-mail: {linda,hansb,ad}@cs.uu.nl

## Abstract

For many real-life Bayesian networks, common knowledge dictates that the output established for the main variable of interest increases with higher values for the observable variables. We define two concepts of monotonicity to capture this type of knowledge. We say that a network is *isotone in distribution* if the probability distribution computed for the output variable given specific observations is stochastically dominated by any such distribution given higher-ordered observations; a network is *isotone in mode* if a probability distribution given higher observations has a higher mode. We show that establishing whether a network exhibits any of these properties of monotonicity is $\text{coNP}^{\text{PP}}$-complete in general, and remains coNP-complete for polytrees. We present an approximate algorithm for deciding whether a network is monotone in distribution and illustrate its application to a real-life network in oncology.

## 1 INTRODUCTION

In most real-life problems, the variables of importance have different roles. Often, a number of observable input variables are distinguished and a single output variable. In a medical diagnostic application, for example, the observable variables capture the findings from different diagnostic tests and the output variable models the possible diseases. Multiple input variables and a single output variable in fact are typically found in any type of classification problem.

For many classification problems, common knowledge dictates that the relation between the output variable and the observable input variables is isotone in the sense that higher values for the input variables should give rise to a higher-ordered output for the main variable of interest. In a medical diagnostic application, for example, observing more severe symptoms and signs should result in a more severe disease being the most likely value of the diagnostic variable. Another example pertains to the domain of loan acceptance where an applicant who scores at least as good on all acceptance criteria as another applicant, should have the higher probability of being accepted. If such knowledge is common sense, then a model that does not exhibit the associated monotonicity properties, will not easily be accepted.

Since monotonicity properties are commonly found in real-life application domains, many modelling techniques have been adapted to capture such properties. Monotonicity has been investigated, for example, for neural networks [1], for decision lists [2], and for classification trees [8], while isotonic regression [9] deals with regression problems with monotonicity constraints. For classification trees, for example, the problem of deciding whether or not a given tree is monotone can be solved in polynomial time. Moreover, efficient learning algorithms have been designed that are guaranteed to result in monotone classification trees [8].

A Bayesian network may also not exhibit the monotonicity properties from its domain of application. In this paper, we introduce two concepts of monotonicity for Bayesian networks. We say that a network is *isotone in distribution* if the probability distribution computed for the output variable given specific observations is stochastically dominated by any such distribution given higher-ordered observations. We further say that the network is *isotone in mode* if the probability distribution computed for the output variable given specific observations has a higher mode than any such distribution given lower-ordered observations. Although the two types of monotonicity are closely related, they capture different properties of a Bayesian network. The first type of monotonicity is more useful, for example, in the context of decision problems where the probability distribution over the output variable is used for further computations; the second type of monotonicity is more useful in the context of problems where the most likely value of the output variable is returned.

For both types of monotonicity, we show that the problem of deciding whether it holds for a given Bayesian network, is complete in general for the complexity class $\text{coNP}^{\text{PP}}$.

The problem of verifying monotonicity thus appears to be highly intractable, and in fact remains so for polytrees. Given this unfavourable complexity, we provide an approximate algorithm for deciding whether a given network is monotone in distribution. Whenever the algorithm indicates that a network is monotone, then it is guaranteed to be so. The algorithm further shows an anytime property: the more time it is granted, the more likely it is to decide whether or not a network is monotone. We demonstrate the application of our algorithm to a real-life network in oncology and argue that it served to identify violation of one of the monotonicity properties from the network's domain.

The present paper is organised as follows. In Section 2, we provide some preliminaries on Bayesian networks and introduce our notational conventions. Our two concepts of monotonicity are introduced in Section 3. In Section 4, we establish the computational complexity of the problem of deciding whether a given network is monotone for both concepts of monotonicity. In Section 5, we present an approximate algorithm for deciding whether or not a given network is monotone in distribution. The paper concludes with some directions for further research in Section 6.

## 2 BAYESIAN NETWORKS

A *Bayesian network* is a representation of a joint probability distribution over a set of stochastic variables [5]. Before briefly reviewing the concept of Bayesian network, we introduce some notational conventions. Stochastic variables are denoted by capital letters. Each variable $V$ can adopt one of a set $\Omega(V)$ of discrete values; we assume that there exists a total ordering $\leq$ on the set $\Omega(V)$. For a binary variable $V$ with the values $v$ and $\bar{v}$ more specifically, we assume that $\bar{v} \leq v$. For any set of variables $S$, we use $\Omega(S) = \times_{V \in S}\, \Omega(V)$ to denote the set of all joint value assignments to $S$; the set $\Omega(\varnothing)$ is defined to include the single element *true*. The total orderings $\leq$ on the sets of values $\Omega(V)$ for the separate variables induce a partial ordering $\preceq$ on the set $\Omega(S)$ of joint value assignments. In mathematical formulas, we will often write $S$ to express that the formula holds for all value assignments to $S$.

A Bayesian network now is a tuple $\mathcal{B} = (G, \Gamma)$ where $G = (V(G), A(G))$ is a directed acyclic graph and $\Gamma$ is a set of conditional probability distributions. In the digraph $G$, each vertex $V \in V(G)$ models a stochastic variable. We assume that the set $V(G)$ is partitioned into three mutually exclusive subsets $X(G)$, $I(G)$ and $C(G)$. The set $X(G) = \{X_1, \ldots, X_n\}$, $n \geq 1$, includes the *observable variables*, that is, the variables for which a value can be established by observation; the set $C(G) = \{C\}$ includes the single *output variable* of the network. For the variables of the set $I(G) = V(G) \setminus (X(G) \cup C(G))$, no value can be observed; these variables are called *intermediate*. The set $A(G)$ of arcs of $G$ captures probabilistic independence: for a topological sort of the digraph $G$, that is, for an ordering $V_1, \ldots, V_m$, $m \geq 1$, of its variables with $i < j$ for every arc $V_i \to V_j \in A(G)$, we have that any variable $V_i$ is independent of the preceding variables $V_1, \ldots, V_{i-1}$ given its parents $\pi(V_i)$. Associated with the digraph $G$ is a set $\Gamma$ of probability distributions: for each variable $V$ are specified the conditional distributions $\Pr(V \mid \pi(V))$ that describe the influence of the various assignments to the variable's parents $\pi(V)$ on the probabilities of the values of $V$ itself.

## 3 CONCEPTS OF MONOTONICITY

Given a joint value assignment to its set of observable variables, a Bayesian network basically serves to compute an output for its main variable of interest. The network can thus be looked upon as a function $f$ that, for each joint value assignment $x$, returns an output $f(x)$. We distinguish between two types of output function for a network.

Let $\mathcal{B}$ be a Bayesian network and let $\Pr$ be the joint probability distribution defined by $\mathcal{B}$. Let $C$ be the output variable of the network and let $X(G)$ be its set of observable variables. The *distribution output function* for $\mathcal{B}$ is a function $f_{\Pr}\colon \Omega(X(G)) \to \mathcal{P}r(C)$, where $\mathcal{P}r(C)$ denotes the set of all possible probability distributions over $C$; the function yields, for each joint value assignment $x$ to $X(G)$, the posterior distribution $\Pr(C \mid x)$ given $x$ over $C$, that is, it is defined as

$$f_{\Pr}(x) = \Pr(C \mid x)$$

for all $x \in \Omega(X(G))$. The *mode output function* for the network is the function $f_{\top}\colon \Omega(X(G)) \to \Omega(C)$ that yields, for each value assignment $x$ to $X(G)$, the value

$$f_{\top}(x) = \top(C \mid x)$$

where $\top(C \mid x)$ denotes the mode of the posterior probability distribution $\Pr(C \mid x)$; if the distribution $\Pr(C \mid x)$ is multimodal, that is, if there are multiple values of $C$ with equal posterior probability, then $\top(C \mid x)$ is defined as the mode that is lowest in the ordering $\leq$ on $\Omega(C)$.

Building upon the mode output function, we now define the concept of monotonicity in mode for a Bayesian network.

**Definition 1** *A Bayesian network $\mathcal{B} = (G, \Gamma)$ is said to be* isotone in mode *for the variables $X(G)$ if*

$$x \preceq x' \;\to\; f_{\top}(x) \leq f_{\top}(x')$$

*for all value assignments $x, x' \in \Omega(X(G))$; if*

$$x \preceq x' \;\to\; f_{\top}(x) \geq f_{\top}(x')$$

*for all $x, x' \in \Omega(X(G))$, then the network is said to be* antitone in mode *for $X(G)$.*

From the definition we have that a Bayesian network is isotone in mode if entering a higher-ordered value assignment to the observable variables, cannot result in a lower-ordered output value for the main variable of interest.

In addition to the concept of monotonicity in mode, we define another concept of monotonicity that builds upon the distribution output function for a Bayesian network. This concept is defined in terms of stochastic dominance. For a probability distribution $\Pr(V)$ over a stochastic variable $V$, the cumulative distribution function $F_{\Pr}$ is defined by $F_{\Pr}(v) = \Pr(V \leq v)$, for all $v \in \Omega(V)$. For two distributions $\Pr(V)$ and $\Pr'(V)$ over $V$, associated with $F_{\Pr}(V)$ and $F_{\Pr'}(V)$, respectively, we say that $\Pr'(V)$ is *stochastically dominant* over $\Pr(V)$, denoted $\Pr(V) \leq \Pr'(V)$, if $F_{\Pr'}(v) \leq F_{\Pr}(v)$, for all $v \in \Omega(V)$. We now define the concept of monotonicity in distribution.

**Definition 2** *A Bayesian network $\mathcal{B} = (G, \Gamma)$ is said to be* isotone in distribution *for the variables $X(G)$ if*

$$x \preceq x' \rightarrow f_{\Pr}(x) \leq f_{\Pr}(x')$$

*for all value assignments $x, x' \in \Omega(X(G))$; if*

$$x \preceq x' \rightarrow f_{\Pr}(x) \geq f_{\Pr}(x')$$

*for all $x, x' \in \Omega(X(G))$, then the network is said to be* antitone in distribution *for $X(G)$.*

From the definition we have that a Bayesian network is isotone in distribution if entering a higher-ordered value assignment to the observable variables, cannot make higher-ordered values of the output variable less likely.

Although the two concepts of monotonicity are closely related, they model different properties of a Bayesian network. An example serves to show that monotonicity in distribution does not imply monotonicity in mode in general. We consider to this end an output variable $C$ with the three possible values $c_1 \leq c_2 \leq c_3$. Suppose that the following posterior distributions over $C$ are computed, given the two value assignments $x$ and $x'$ with $x \preceq x'$:

$$\begin{array}{ll} \Pr(c_1 \mid x) = 0.25 & \Pr(c_1 \mid x') = 0 \\ \Pr(c_2 \mid x) = 0.35 & \Pr(c_2 \mid x') = 0.55 \\ \Pr(c_3 \mid x) = 0.4 & \Pr(c_3 \mid x') = 0.45 \end{array}$$

From the associated cumulative distributions, we find that the probability distribution $\Pr(C \mid x')$ is stochastically dominant over $\Pr(C \mid x)$. The network thus is isotone in distribution. We further observe that the mode of the distribution $\Pr(C \mid x')$ equals $c_2$, while the mode of $\Pr(C \mid x)$ equals $c_3$. We thus have that $\top(C \mid x') \leq \top(C \mid x)$, from which we conclude that the network is not isotone in mode. By reversing the roles of the assignments $x$ and $x'$ in the argument, it is readily seen that monotonicity in mode also does not imply monotonicity in distribution.

Even for a binary output variable do the two concepts of monotonicity not coincide. For a binary output variable $C$, we have that monotonicity in distribution *does* imply monotonicity in mode. Suppose that the network under consideration is isotone in distribution. For any two value assignments $x$ and $x'$ with $x \preceq x'$, we then have that $f_{\Pr}(x) \leq f_{\Pr}(x')$, from which we find that $\Pr(\bar{c} \mid x') \leq \Pr(\bar{c} \mid x)$ and, hence, $\Pr(c \mid x) \leq \Pr(c \mid x')$. We conclude that the network is isotone in mode. The reverse property does not hold, however. Suppose that, from the network, for the two value assignments $x$ and $x'$, the following posterior probability distributions over $C$ are computed:

$$\begin{array}{ll} \Pr(\bar{c} \mid x) = 0.6 & \Pr(\bar{c} \mid x') = 0.9 \\ \Pr(c \mid x) = 0.4 & \Pr(c \mid x') = 0.1 \end{array}$$

The mode of both distributions equals $\bar{c}$. We thus have that $\top(C \mid x) \leq \top(C \mid x')$ and we conclude that the network is isotone in mode. From the associated cumulative distributions, however, we observe that the probability distribution $\Pr(C \mid x)$ is stochastically dominant over $\Pr(C \mid x')$. The network therefore is not isotone in distribution.

## 4 COMPUTATIONAL COMPLEXITY

To establish the computational complexity of the problems of deciding whether a Bayesian network exhibits the properties of monotonicity in distribution and monotonicity in mode, we formulate them as decision problems.

**Definition 3** *Let $\mathcal{B} = (G, \Gamma)$ be a Bayesian network where $\Gamma$ is composed of rational probabilities, and let $\Pr$ be its joint probability distribution. Let $X(G)$ be the set of observable variables of $\mathcal{B}$ and let $C$ be its output variable.*

*a. The* MIM *problem is the problem of deciding whether for all value assignments $x$ and $x'$ to $X(G)$ with $x \preceq x'$, it holds that $\top(C \mid x) \leq \top(C \mid x')$.*

*b. The* MID *problem is the problem of deciding whether for all value assignments $x$ and $x'$ to $X(G)$ with $x \preceq x'$, it holds that $\Pr(C \mid x) \leq \Pr(C \mid x')$.*

*The* NOT-MIM *and* NOT-MID *problems are the complements of the* MIM *and* MID *problems, respectively.*

In the remainder of this section, we address the computation complexity of the problems defined above. More specifically, we show that the NOT-MIM problem is complete for the complexity class $\text{NP}^{\text{PP}}$, from which we have that the MIM problem is $\text{coNP}^{\text{PP}}$-complete. A similar complexity result then follows directly for the MID problem.

To establish the intractability of the NOT-MIM problem, we use a reduction from a decision version of the well-known MAP problem. Let $\mathcal{B} = (G, \Gamma)$ be a Bayesian network with rational probabilities and let $\Pr$ be its joint probability distribution. Let $E(G)$ be the subset of observed variables of $\mathcal{B}$ and let $e$ be the available evidence; let $M(G)$ be the set of variables for which we want to find a joint value assignment of maximum probability in the presence of $e$, where $E(G) \cap M(G) = \varnothing$. Let $p \in [0,1]$ be a rational number. The MAP problem now is the problem of deciding whether there exists a joint value assignment $m$ to $M(G)$ such that $\Pr(me) > p$. The MAP problem was shown to be complete for the complexity class $\mathrm{NP}^{\mathrm{PP}}$ [7]; the problem further was shown to remain NP-complete for polytrees.

The proof of hardness for the class $\mathrm{NP}^{\mathrm{PP}}$ of the MAP problem [7], builds upon the construction of a network in which the MAP variables $M(G)$ do not have any incoming arcs and have specified a uniform prior probability distribution. In the proof therefore, only value assignments $m$ to $M(G)$ are considered for which $\Pr(m)$ is the same rational number. The proof further uses a singleton set $E(G)$ with a binary evidence variable $E$. Based upon these observations, it is readily seen that also the COND-MAP problem is $\mathrm{NP}^{\mathrm{PP}}$-complete and remains NP-complete for polytrees. Let $\mathcal{B}$, Pr, $M(G)$, and $p$ be as before, and let $E$ be a binary variable for which the evidence $e$ has been observed. Then, the COND-MAP problem is the problem of deciding whether there exists a joint value assignment $m$ to $M(G)$ such that $\Pr(e \mid m) > p$.

We now show that the NOT-MIM problem is $\mathrm{NP}^{\mathrm{PP}}$-hard. Our proof is based upon the following reduction. Suppose that we have a Bayesian network $\mathcal{B} = (G, \Gamma)$ as before and a rational number $p \in [0,1]$. Further suppose that we have a set $X(G)$ of observable variables for which no evidence has been obtained as yet, and that we have observed a single piece of evidence $e$ for the binary variable $E \notin X(G)$. We now build a new Bayesian network $\mathcal{B}' = (G', \Gamma')$ as follows. To obtain the digraph $G'$, we add three new, binary variables $A$, $B$ and $C$ to the digraph $G$ of $\mathcal{B}$. For the sets of values $\Omega(V)$ for all variables $V \in V(G) \cap V(G')$, we adopt the same ordering as given for the network $\mathcal{B}$; for the new variables, we assume the general ordering for binary variables indicated before. In addition, we add to the digraph $G'$, the three arcs $E \to A$, $A \to C$ and $B \to C$. Figure 1 illustrates the basic idea.

To obtain the set $\Gamma'$ of probability distributions for $\mathcal{B}'$, we add probability distributions for the new variables to the set $\Gamma$. For the variable $B$, we specify a uniform prior distribution. For the variables $A$ and $C$, we add the distributions

$$\Pr(a \mid E) = \begin{cases} 1 & \text{if } E = e \\ \dfrac{\frac{1}{2} - p}{1 - p} & \text{otherwise} \end{cases}$$

and

$$\Pr(c \mid AB) = \begin{cases} 1 & \text{if } A = a \text{ and } B = \bar{b} \\ 0 & \text{otherwise} \end{cases}$$

respectively. The set $X(G') = X(G) \cup \{B\}$ now is taken for the set of observable variables of the new network $\mathcal{B}'$; $C$ is taken for its output variable.

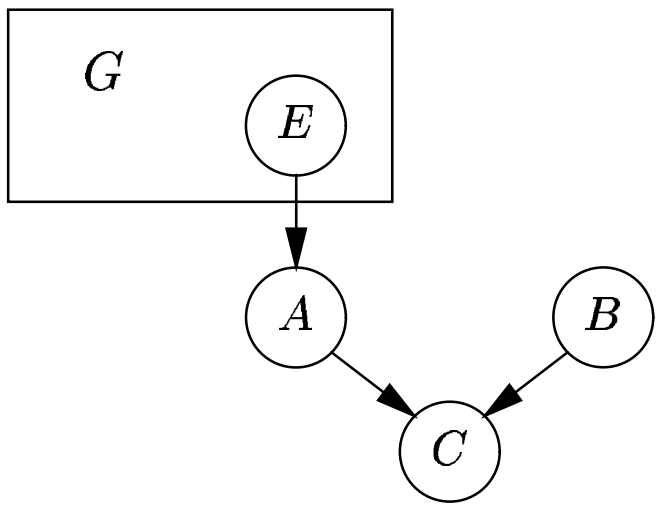

Figure 1: The construction of the digraph of $\mathcal{B}'$.

**Lemma 1** *Let $\mathcal{B} = (G, \Gamma)$ be a Bayesian network with the observable variables $X(G)$ as before and let $\mathcal{B}' = (G', \Gamma')$ be the Bayesian network with the observable variables $X(G')$ that is constructed from $\mathcal{B}$ as indicated above. Then, there is a value assignment $x''$ to $X(G)$ with $\Pr(e \mid x'') > p$ if and only if there are value assignments $x$ and $x'$ to $X(G')$ with $x \preceq x'$ and $\top(C \mid x) > \top(C \mid x')$.*

**Proof**. Suppose that there is a value assignment $x''$ to $X(G)$ with $\Pr(e \mid x'') > p$. Now, let $x$ be the value assignment to $X(G') = X(G) \cup \{B\}$ that is obtained by assigning to each variable $X \in X(G)$ its value from $x''$ and by assigning the value $\bar{b}$ to $B$; let $x'$ be the assignment to $X(G')$ that is obtained similarly to $x$, yet with the value $b$ for $B$. We then have that $\top(C \mid x) = c$. To prove this observation, we begin by noting that $\Pr(E \mid x) = \Pr(E \mid x'')$ since $E$ is independent of $B$ given $X(G)$. From $\Pr(e \mid x'') > p$, we thus have that $\Pr(e \mid x) > p$. For the probability $\Pr(a \mid x)$, we further find that

$$\Pr(a \mid x) = \Pr(a \mid e) \cdot \Pr(e \mid x) + \Pr(a \mid \bar{e}) \cdot \Pr(\bar{e} \mid x) =$$
$$> p + (1-p) \cdot \frac{\frac{1}{2} - p}{1 - p} = \tfrac{1}{2}$$

Since the variable $B$ has been assigned the value $\bar{b}$ in $x$, we find that $\Pr(c \mid x) = \Pr(a \mid x) > \frac{1}{2}$, from which we conclude that $\top(C \mid x) = c$. Since the variable $B$ has the value $b$ in the assignment $x'$, we have that $\Pr(c \mid x') = 0$ regardless of the other values in $x'$. We conclude that $\top(C \mid x') = \bar{c}$. From the above observations we have that $\top(C \mid x) = c > \top(C \mid x') = \bar{c}$; from the construction of the assignments $x$ and $x'$ we further have that $x \preceq x'$.

Now, suppose that there are value assignments $x$ and $x'$ to $X(G')$ with $x \preceq x'$ and $\top(C \mid x) > \top(C \mid x')$. Since the variable $C$ is binary, we then have that $\top(C \mid x) = c$ and $\top(C \mid x') = \bar{c}$. From $\top(C \mid x) = c$, we conclude that $\Pr(c \mid x) > \frac{1}{2}$. Now, let $x''$ be the value assignment to $X(G)$ that is obtained by assigning to each variable $X \in X(G)$, its value from $x$. Let $\Pr(e \mid x'') = q$ for some $q \in [0,1]$. We then have that $\Pr(e \mid x) = q$ and, hence, that $\Pr(a \mid x) = q + (1-q) \cdot \frac{\frac{1}{2}-p}{1-p}$. Since $B$ has the value $\bar{b}$ in $x$, we have that $\Pr(c \mid x) = \Pr(a \mid x) = q + (1-q) \cdot \frac{\frac{1}{2}-p}{1-p}$. From $\Pr(c \mid x) > \frac{1}{2}$, we now find that $q > p$. We conclude that $\Pr(e \mid x'') > p$. □

Based upon the computational complexity of the COND-MAP problem and the result stated in Lemma 1, we have that the NOT-MIM problem is $NP^{PP}$-hard. We further established membership of the problem in $NP^{PP}$, the proof of which is omitted for reasons of space. We conclude that the NOT-MIM problem is $NP^{PP}$-complete. We further observe that, if the digraph $G$ of the original network is a polytree, then the digraph $G'$ used in our reduction is also a polytree. From the computational complexity of the COND-MAP problem for polytrees, we thus have that the NOT-MIM problem also remains NP-complete for polytrees.

To conclude, building upon the definition of $coNP^{PP}$-completeness and the completeness of the NOT-MIM problem for the class $NP^{PP}$, we now state for our main complexity result that the MIM problem is $coNP^{PP}$-complete and remains coNP-complete for polytrees.

## 5 APPROXIMATING MONOTONICITY IN DISTRIBUTION

From our definitions, we have that establishing whether or not a Bayesian network is monotone in distribution amounts to verifying that entering a higher-ordered value assignment to the observable variables results in a stochastically dominant probability distribution over the main variable of interest. In Section 5.2, we present an approximate algorithm for verifying monotonicity in distribution. The algorithm builds on the concept of qualitative influence, which is reviewed in Section 5.1. In Section 5.3, we demonstrate the application of our algorithm to a real-life Bayesian network in oncology.

### 5.1 QUALITATIVE INFLUENCE

The concept of qualitative influence has been designed to capture the probabilistic influences between stochastic variables in a qualitative way [10]; the concept is commonly used in qualitative probabilistic networks.

A qualitative influence between two stochastic variables expresses how observing a value for the one variable affects the probability distribution over the other variable. A positive qualitative influence of the variable $V$ on the variable $W$, for example, expresses that observing a higher value for $V$ makes higher values for $W$ more likely, regardless of any other direct influences on $W$. More formally, $V$ has a *positive qualitative influence* on $W$ if, for all values $w$ of $W$ and all values $v, v'$ of $V$, with $v \leq v'$, we have that

$$F_{\Pr}(w \mid v's) \leq F_{\Pr}(w \mid vs)$$

for any joint value assignment $s$ to the set $S = \pi(W) \setminus \{V\}$ of parents of $W$ other than $V$; the set $S$ is termed the *context set* for the influence. We say that the influence is associated with the *sign* '$+$'. A *negative qualitative influence*, associated with the sign '$-$', and a *zero qualitative influence*, associated with a '0', are defined analogously, replacing $\leq$ in the above formula by $\geq$ and $=$, respectively. For a positive, negative or zero qualitative influence of the variable $V$ on the variable $W$, we have that the difference

$$F_{\Pr}(w \mid vs) - F_{\Pr}(w \mid v's)$$

has the same sign for *all* value assignments to the set $S$. This sign then is guaranteed to hold for any (fixed) probability distribution over the context variables. If the influence of $V$ on $W$ is positive given one value assignment to $S$ and negative given another assignment, then the influence is called *non-monotone* and is associated with a '?'.

The set of all qualitative influences between the variables of a Bayesian network exhibits some important properties [10]. The property of symmetry states that, if the network includes an influence of sign $\delta$ of $V$ on $W$, $\delta \in \{+, -, 0, ?\}$, then it also includes an influence of $W$ on $V$ of sign $\delta$. The transitivity property asserts that the qualitative influences along a trail that specifies at most one incoming arc for each variable, combine into a net influence whose sign is defined by the $\otimes$-operator from Table 1; a net influence of $V$ on $W$ along a given trail $t$ that is composed of the variables $V(t)$, has for its context the set $S$ with

$$S = \left( \bigcup_{U \in V(t) \setminus \{V\}} \pi(U) \right) \setminus V(t)$$

To conclude, the property of composition asserts that multiple influences between two variables along parallel trails combine into a net influence whose sign is defined by the $\oplus$-operator; the context set for the combined influence is defined as above. The three properties with each other provide for establishing the sign of an indirect influence between any two variables in a network.

For computing the signs of indirect qualitative influences from the signs of the direct influences in a Bayesian network, an efficient algorithm is available [3]. This algorithm provides for establishing the qualitative effect of an observation upon the probability distributions for the other variables. It is based on the idea of propagating and combining signs, and builds upon the properties of symmetry, transitivity and composition of qualitative influences. The joint effect of multiple observations on a variable of interest can be computed as the $\oplus$-sum of the effects of the separate observations on this variable's probability distribution. The algorithm has a runtime complexity that is polynomial in the number of variables in a network.

| $\otimes$ | $+$ | $-$ | 0 | ? |
|---|---|---|---|---|
| $+$ | $+$ | $-$ | 0 | ? |
| $-$ | $-$ | $+$ | 0 | ? |
| 0 | 0 | 0 | 0 | 0 |
| ? | ? | ? | 0 | ? |

| $\oplus$ | $+$ | $-$ | 0 | ? |
|---|---|---|---|---|
| $+$ | $+$ | ? | $+$ | ? |
| $-$ | ? | $-$ | $-$ | ? |
| 0 | $+$ | $-$ | 0 | ? |
| ? | ? | ? | ? | ? |

Table 1: The $\otimes$- and $\oplus$-operators for combining signs.

### 5.2 VERIFYING MONOTONICITY

For verifying monotonicity of a Bayesian network, we first study the relation between a single observable variable and the main variable of interest. From the definitions of monotonicity in distribution and of the concept of qualitative influence, we now have that if the observable variable exerts a positive influence on the output variable, then the network is isotone in distribution for the input variable under study.

**Lemma 2** *Let $\mathcal{B} = (G, \Gamma)$ be a Bayesian network with the observable variable $X$ and the output variable $C$ as before.*

a. *If $X$ has a positive qualitative influence on $C$, then $\mathcal{B}$ is isotone in distribution for $X$.*

b. *If $X$ has a negative qualitative influence on $C$, then $\mathcal{B}$ is antitone in distribution for $X$.*

**Proof**. We prove the first property stated in the lemma only; similar observations hold for the second property. We suppose that the variable $X$ exerts a positive overall qualitative influence on $C$. By definition, we then have that

$$x \leq x' \rightarrow F_{\Pr}(c \mid x's) \leq F_{\Pr}(c \mid xs)$$

for all values $x, x' \in \Omega(X)$, for all assignments $s$ to the context set $S$ of the influence, and for all values $c \in \Omega(C)$. Since the sign for the influence holds for any probability distribution over $S$, we have that

$$x \leq x' \rightarrow F_{\Pr}(c \mid x') \leq F_{\Pr}(c \mid x)$$

from which we find that

$$x \leq x' \rightarrow f_{\Pr}(x) \leq f_{\Pr}(x')$$

The network thus is isotone in distribution for $X$. □

An approximate algorithm for establishing whether or not a Bayesian network $\mathcal{B} = (G, \Gamma)$ with a single observable variable $X$ and an output variable $C$ is monotone in distribution for $X$, now amounts to the following:

a. Establish, for each arc in the digraph $G$ of the network, the sign of the associated influence, by inspection of the network's set $\Gamma$ of probability distributions.

b. From the established qualitative influences, compute the sign of the overall influence of $X$ on $C$.

c. For the computed sign $\delta$ of the influence of $X$ on $C$:

- if $\delta = +$, then the network is isotone in distribution for $X$;
- if $\delta = -$, then the network is antitone in distribution for $X$;
- if $\delta = 0$, then the network is both isotone and antitone in distribution for $X$;
- if $\delta = ?$, then neither isotonicity nor antitonicity can be established for $X$.

From Lemma 2, we have that, if the algorithm returns isotonicity or antitonicity, then its outcome is indeed correct. If the algorithm does not return any monotonicity and hence is inconclusive, then the network in fact may or may not be monotone for the variable under study. The algorithm, however, never returns an incorrect outcome. We note that the algorithm has a runtime complexity that is polynomial in the number of variables in the network.

While a positive qualitative influence of the observable variable $X$ on the output variable implies isotonicity in distribution, the reverse property does not hold: if a Bayesian network is isotone in distribution, then the observable variable need not exert a positive overall qualitative influence on the main variable of interest. To support this observation, we suppose that the network exhibits the property of isotonicity in distribution for its input variable $X$. By definition, we then have that

$$x \leq x' \rightarrow f_{\Pr}(x) \leq f_{\Pr}(x')$$

from which we find that

$$x \leq x' \rightarrow F_{\Pr}(c \mid x') \leq F_{\Pr}(c \mid x)$$

for all values $x, x' \in \Omega(X)$ and for all values $c \in \Omega(C)$. The latter property now holds for the prior probability distribution over the context set $S$ of the influence of $X$ on $C$, yet may not hold for all possible distributions over $S$. The variable $X$ may therefore have a non-monotone influence on $C$. As an example, we consider the Bayesian network from Figure 2; in the network, $X_1$ is the only observable variable, $C$ is the output variable, and $X_2$ and $Y$ are intermediate variables. From the probability distributions specified for the variable $C$, we find that the qualitative influence of $X_1$ on $C$ is non-monotone: if the value $y$ would be observed, the influence would be negative; if the value $\bar{y}$ would be observed, the influence would be positive. We note, however, that the intermediate variables $Y$ and $X_2$ cannot be observed. With the specified probabilities, we have that $\Pr(y) = 0.32$, from which we find that $\Pr(c \mid x_1) = 0.87$ and $\Pr(c \mid \bar{x}_1) = 0.56$. Building upon the non-observability of both $X_2$ and $Y$, therefore, we find that the influence of $X_1$ on $C$ actually is positive. We conclude that, while the network is isotone for $X_1$, the qualitative influence of $X_1$ on $C$ has a non-monotone sign.

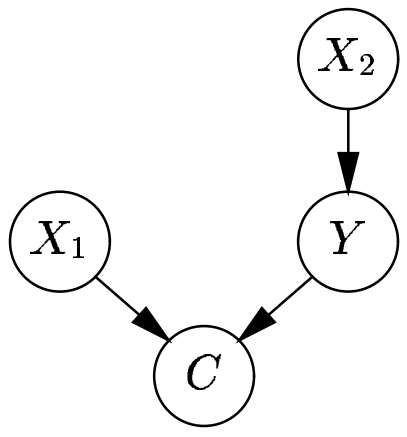

$\Pr(x_1) = 0.75$

$\Pr(x_2) = 0.9$

$\Pr(y \mid x_2) = 0.3$
$\Pr(y \mid \bar{x}_2) = 0.5$

$\Pr(c \mid x_1 y) = 0.8$
$\Pr(c \mid x_1 \bar{y}) = 0.9$
$\Pr(c \mid \bar{x}_1 y) = 0.9$
$\Pr(c \mid \bar{x}_1 \bar{y}) = 0.4$

Figure 2: An example Bayesian network.

We now consider a Bayesian network with multiple observable variables and study the relation between the main variable of interest and these input variables. We find that, if each input variable separately exhibits a positive qualitative influence on the output variable, then the network is isotone in distribution for the entire set of input variables.

**Theorem 1** *Let $\mathcal{B} = (G, \Gamma)$ be a Bayesian network with the set $X(G)$ of observable variables and the output variable $C$ as before.*

*a. If, for all $X_i \in X(G)$, $X_i$ has a positive qualitative influence on $C$, then $\mathcal{B}$ is isotone in distribution for $X(G)$.*

*b. If, for all $X_i \in X(G)$, $X_i$ has a negative qualitative influence on $C$, then $\mathcal{B}$ is antitone in distribution for $X(G)$.*

**Proof**. We prove the first property stated in the theorem only; similar observations hold for the second property. We suppose that each variable $X_i \in X(G)$ exerts a positive overall influence on the output variable $C$, that is, for each variable $X_i$, we have that

$$x_i \leq x_i' \rightarrow F_{\mathrm{Pr}}(c \mid x_i' s) \leq F_{\mathrm{Pr}}(c \mid x_i s)$$

for all values $x_i, x_i' \in \Omega(X_i)$, for all value assignments $s$ to the context set $S$, and for all values $c \in \Omega(C)$. Since the inequalities moreover hold for any probability distribution over $S$, we have that

$$x_i x^- \preceq x_i' x^- \rightarrow F_{\mathrm{Pr}}(c \mid x_i' x^-) \leq F_{\mathrm{Pr}}(c \mid x_i x^-)$$

for any value assignment $x^-$ to the set of variables $X(G) \setminus \{X_i\}$. We thus find that

$$x \preceq x' \rightarrow F_{\mathrm{Pr}}(c \mid x') \leq F_{\mathrm{Pr}}(c \mid x)$$

for all assignments $x, x' \in \Omega(X(G))$ and all values $c \in \Omega(C)$. We conclude that the network is isotone in distribution for $X(G)$. □

We note that the previous theorem provides for identifying from a set of observable variables, a subset of variables for which the network under study is isotone and a subset of variables for which it is antitone. The theorem in addition provides for identifying the variables which forestall a conclusion with respect to monotonicity based upon just their qualitative relation with the main variable of interest.

While positive qualitative influences of all observable variables separately on the output variable imply isotonicity in distribution of a network, the reverse property does not hold. As an example, we consider again the network from Figure 2. We now assume that both $X_1$ and $X_2$ are observable; $C$ again is the output variable and $Y$ is the only intermediate variable. From the probability distributions specified for the variable $C$, we again find that the qualitative influence of $X_1$ on $C$ is non-monotone. In view of the two possible observations for the variable $X_2$, we now find that the probability of $y$ ranges between $0.3$ and $0.5$. With these bounds, we find that $\Pr(c \mid x_1) \in [0.67, 0.69]$ and $\Pr(c \mid \bar{x}_1) \in [0.47, 0.57]$. Building upon the non-observability of $Y$, therefore, we conclude that for all possible distributions over $Y$, the influence of $X_1$ on $C$ is positive. We conclude that, while the network is isotone for $X_1$, the qualitative influence of $X_1$ on $C$ is non-monotone.

Building upon the properties stated in Theorem 1, we can readily extend our approximate algorithm for establishing whether or not a Bayesian network with a single observable variable is monotone in distribution, to apply to multiple input variables. For the extended algorithm, we again have that, if the algorithm returns isotonicity or antitonicity, then its outcome is correct. The previous example further illustrates that, if the outcome for a given network is inconclusive, then establishing bounds on the probability distributions for the intermediate variables may help the algorithm to reach a conclusive outcome. Such bounds can be computed using an algorithm available for this purpose from Liu and Wellman [6]. This algorithm has an exponential runtime complexity, yet exhibits an anytime property in that the more time it is granted, the tighter the computed bounds are. By including this algorithm into our algorithm for verifying monotonicity, our algorithm inherits the anytime property: the more time it is granted, the more likely it is to decide whether or not a given network is monotone.

### 5.3 AN EXAMPLE

We applied our algorithm for verifying monotonicity to a real-life Bayesian network in the field of cancer of the oesophagus. The OESOCA network provides for establishing the stage of a patient's oesophageal cancer, based upon the results of a number of diagnostic tests. The network has been constructed with the help of gastroenterologists from the Netherlands Cancer Institute, Antoni van Leeuwenhoekhuis. It was evaluated using the medical records from $156$ real patients with cancer of the oesophagus and was found to have a classification accuracy of some $85\%$. Notwithstanding the good overall performance of the network, the evaluation served to identify a specific class of patients for whom the network established an incorrect stage [4]. From the original fully quantified OESOCA network, we constructed a binary qualitative network for our experiment. Figure 3 shows the resulting network; the signs of the direct qualitative influences are shown over the digraph's arcs; the figure in addition shows the prior probability distributions for the variables involved.

The qualitative OESOCA network has the variable *Stage* for its main diagnostic variable. It further includes 23 variables that serve to model the results of diagnostic tests; these variables are leaves of the network's digraph. Domain knowledge now dictates that the network should be monotone in distribution for 21 of the input variables; for two variables, there appears to be no natural order on their val-ues. Upon applying our algorithm for verifying monotonicity to these 21 observable variables, a conclusive outcome was found for 19 of them. For 90% of the observable variables, therefore, the algorithm correctly concluded that the network is monotone in distribution. For the remaining two variables, the algorithm yielded an inconclusive outcome. Closer examination of these two variables revealed that the expected monotonicity property indeed did not hold. More specifically, we found that violation of this monotonicity property served to explain the poor performance of the network for the previously identified class of patients.

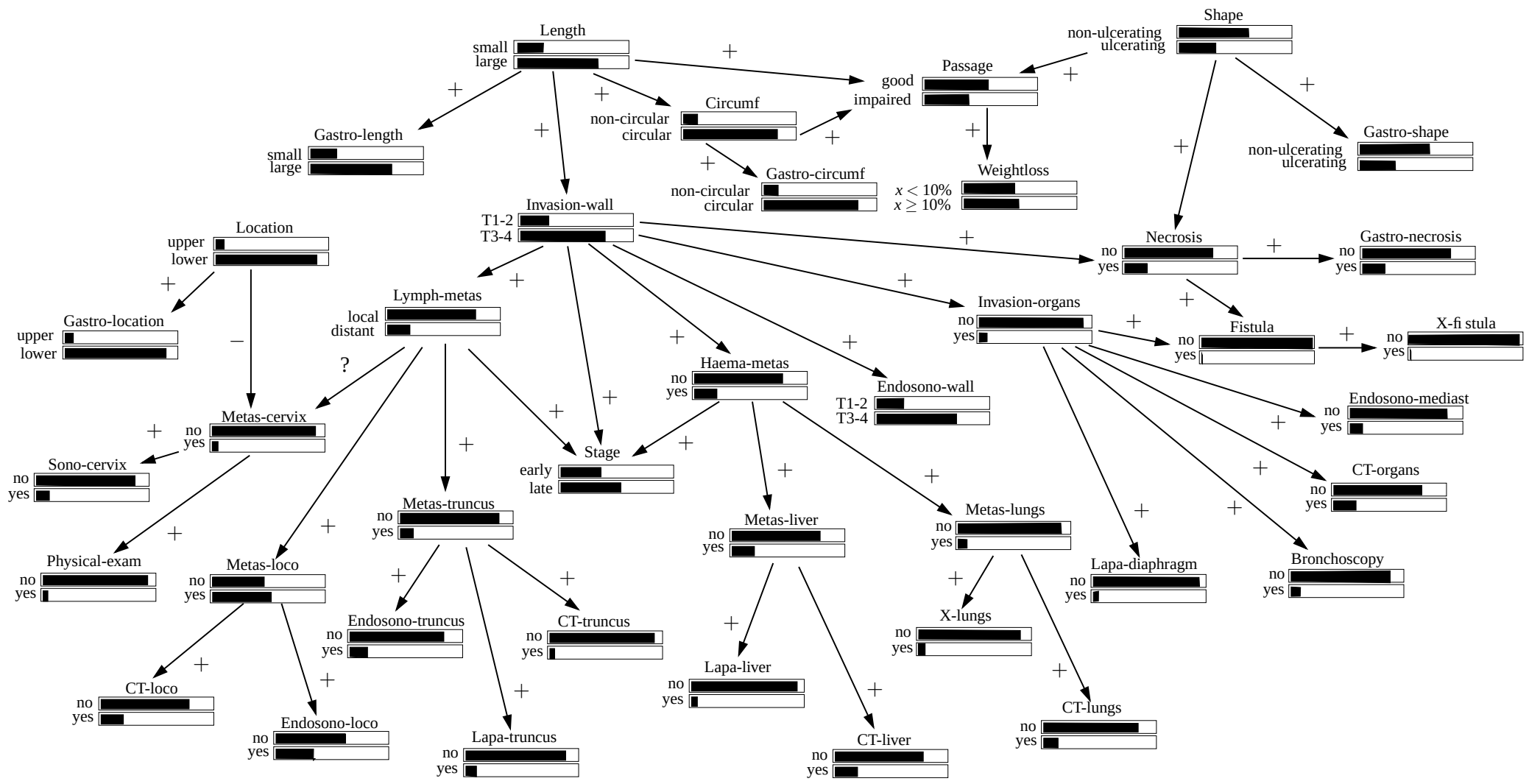

Figure 3: The qualitative OESOCA network.

## 6 CONCLUSIONS

We introduced two new concepts of monotonicity for Bayesian networks and established the computational complexity of verifying whether any of these monotonicity properties holds for a given network. In view of the unfavourable complexity found, we presented an approximate algorithm for verifying monotonicity in distribution. We reported on the application of our algorithm to a real-life Bayesian network. We note that our algorithm for verifying monotonicity in distribution is based upon a decomposition property that allows for verifying monotonicity for each observable variable separately. No such property has been identified as yet for verifying monotonicity in mode. We further note that so far we addressed the problem of verifying monotonicity only. We are currently studying the problem of learning probability distributions from data that are guaranteed to result in a monotone network. In the near future, we hope to present an algorithm for this purpose.

## Acknowledgements

We would like to thank Jan van Leeuwen for sharing with us his insights into the complexity class (co)$NP^{PP}$.